\PassOptionsToPackage{nopatch=footnote}{microtype}
\documentclass[sigconf,nonacm]{acmart}

\usepackage[utf8]{inputenc} %
\usepackage[T1]{fontenc}    %

\usepackage{parskip}        %
\usepackage{url}            %
\usepackage{booktabs}       %
\usepackage{amsfonts}       %
\usepackage{nicefrac}       %
\usepackage{xcolor}         %
\usepackage[dvipsnames]{xcolor} %
\usepackage{graphicx}
\usepackage{animate}        %
\usepackage{subcaption}
\usepackage{tabularx}
\usepackage{makecell}
\usepackage{adjustbox}
\usepackage{setspace}
\newcolumntype{M}[1]{>{\centering\arraybackslash}m{#1}}
\usepackage{float}
\usepackage{tikz}
\usetikzlibrary{positioning,shapes,arrows}
\usepackage{amsmath,amsfonts,bm, bbm,leftindex}
\usepackage{multirow}
\usepackage{comment}
\usepackage{gensymb}
\usepackage{lipsum}
\usetikzlibrary{arrows.meta, positioning, fit}
\usepackage[para]{threeparttable}
\usepackage{tikz}
\usetikzlibrary{tikzmark}
\usepackage{hyperref}
\usepackage[most]{tcolorbox}
\usepackage{fancyvrb}
\usepackage{fvextra}
\usepackage{dashrule}
\usepackage{wrapfig}
\usepackage{float}

\def\eqref#1{equation~\ref{#1}}

\def\1{\bm{1}}

\DeclareMathAlphabet{\mathsfit}{\encodingdefault}{\sfdefault}{m}{sl}
\SetMathAlphabet{\mathsfit}{bold}{\encodingdefault}{\sfdefault}{bx}{n}

\makeatletter
\let\save@mathaccent\mathaccent
\makeatother

\usepackage{pifont}

\usepackage[nameinlink]{cleveref}
\usepackage{placeins}
\crefname{equation}{Eq.}{Eqs.}
\crefname{figure}{Fig.}{Figs.}
\crefname{section}{Sec.}{Sec.}
\crefname{appendix}{App.}{App.}
\crefname{table}{Tab.}{Tabs.}
\crefname{algorithm}{Algo}{Algo}
\crefname{thm}{Thm}{Thm}
\Crefname{thm}{Thm}{Thm}
\crefname{prop}{Prop}{Prop}
\makeatletter
\AddToHook{cmd/appendix/before}{\def\cref@section@alias{appendix}}
\makeatother

\usepackage{makecell}
\usepackage[table]{xcolor}

\newif\ifshowtodos
\showtodostrue     %
\ifshowtodos
  \newcommand{\todo}[1]{\textcolor{red}{[\textit{TODO: #1}]}\xspace}
\else
  \newcommand{\todo}[1]{}
\fi

\definecolor{darkred}{rgb}{0.7, 0.0, 0.0}

\newcommand{\crefnames}[3]{%
  \@for\next:=#1\do{%
    \expandafter\crefname\expandafter{\next}{#2}{#3}%
  }%
}

\usepackage[most]{tcolorbox}
\tcbuselibrary{listings,skins,breakable}

\newtcolorbox{promptbox}[2][]{enhanced,
  breakable,
  colback=gray!5,
  colframe=gray!60,
  fonttitle=\bfseries,
  coltitle=black,
  title={Prompt~\thetcbcounter: #2},
  label={#1},
  boxrule=0.5pt,
  arc=2pt,
  left=6pt,
  right=6pt,
  top=6pt,
  bottom=6pt,
}

\title{Model Optimization for Multi-Camera 3D Detection and Tracking}

\author{Ethan Anderson}
\affiliation{%
  \institution{Clemson University}
  \city{Clemson}
  \state{SC}
  \country{USA}
}

\author{Justin Silva}
\affiliation{%
  \institution{Clemson University}
  \city{Clemson}
  \state{SC}
  \country{USA}
}

\author{Kyle Zheng}
\affiliation{%
  \institution{Clemson University}
  \city{Clemson}
  \state{SC}
  \country{USA}
}

\author{Sameer Pusegaonkar}
\affiliation{%
  \institution{NVIDIA}
  \city{Santa Clara}
  \state{CA}
  \country{USA}
}

\author{Yizhou Wang}
\affiliation{%
  \institution{NVIDIA}
  \city{Santa Clara}
  \state{CA}
  \country{USA}
}

\author{Zheng Tang}
\affiliation{%
  \institution{NVIDIA}
  \city{Santa Clara}
  \state{CA}
  \country{USA}
}

\author{Sujit Biswas}
\affiliation{%
  \institution{NVIDIA}
  \city{Santa Clara}
  \state{CA}
  \country{USA}
}

\begin{document}


\begin{abstract}
Outside-in multi-camera perception is increasingly important in indoor environments, where networks of static cameras must support multi-target tracking under occlusion and heterogeneous viewpoints. We evaluate Sparse4D, a query-based spatiotemporal 3D detection and tracking framework that fuses multi-view features in a shared world frame and propagates sparse object queries via instance memory. We study reduced input frame rates, post-training quantization (INT8 and FP8), transfer to the WILDTRACK benchmark, and Transformer Engine mixed-precision fine-tuning. To better capture identity stability, we report Average Track Duration (AvgTrackDur), which measures identity persistence in seconds. Sparse4D remains stable under moderate FPS reductions, but below 2 FPS, identity association collapses even when detections are stable. Selective quantization of the backbone and neck offers the best speed-accuracy trade-off, while attention-related modules are consistently sensitive to low precision. On WILDTRACK, low-FPS pretraining yields large zero-shot gains over the base checkpoint, while small-scale fine-tuning provides limited additional benefit. Transformer Engine mixed precision reduces latency and improves camera scalability, but can destabilize identity propagation, motivating stability-aware validation.
\end{abstract}

\maketitle

\section{Introduction}
\label{sec::intro}

Multi-camera perception is increasingly essential in indoor environments such as warehouses, retail stores, and hospitals, where networks of static, widely distributed cameras monitor activity, improve operational efficiency, and support safety. Unlike ego-centric autonomous driving pipelines that can exploit ego-motion cues \cite{wang2023exploringobjectcentrictemporalmodeling, park2022timetellnewoutlooks} and often LiDAR priors, these ``outside-in'' deployments must aggregate evidence across heterogeneous fixed cameras with disconnected fields of view, heavy occlusions from shelving or furniture, and substantial lighting variation. These challenges make long-horizon identity consistency difficult, motivating extensive work on multi-camera association, including approaches on re-identification \cite{ye2021deep, zheng2019joint, hsu2021multi}. Achieving reliable \textit{multi-target multi-camera tracking} (MTMC) under these conditions remains challenging.

Spatiotemporal architectures such as Sparse4D~\cite{lin2022sparse4d,lin2023sparse4dv2,lin2023sparse4dv3} have recently shown strong performance by integrating multi-view image features over time using sparse 3D queries. In contrast to pipelines that rely on explicit BEV intermediate representations, Sparse4D performs multi-view aggregation and 3D reasoning directly in a shared world coordinate frame, making it a compelling baseline for large-scale indoor “outside-in” tracking. However, practical deployment constraints are not captured by accuracy alone. Real camera networks may operate at reduced frame rates due to bandwidth and storage limits, production systems impose latency targets, and deployment hardware often benefits from reduced numerical precision via quantization or mixed-precision training. These factors can disproportionately affect temporal association and identity stability, which are central to MTMC applications.

Rather than proposing new modeling components, this paper evaluates an adapted Sparse4D implementation \cite{wang2026unified3dobjectperception} for outside-in fixed-camera deployments under a suite of deployment-driven conditions designed to surface distinct practical trade-offs:

\textbf{Low-FPS inference robustness:} we progressively reduce the effective inference FPS to identify the failure point where tracking and identity stability collapse, providing a concrete bound on how far FPS can be stretched in cases of adding cameras or reducing hardware costs. 

\textbf{Post-training quantization (PTQ):} we evaluate selective INT8 and FP8 PTQ as an inference latency reduction mechanism, enabling higher camera counts and/or lower-cost hardware without retraining.

\textbf{WILDTRACK adaptation and fine-tuning:} we adapt Sparse4D to WILDTRACK \cite{WILDTRACK_Chavdarova2017TheWM} to assess real-world generalization, measuring how a model trained in indoor warehouse conditions transfers to outdoor multi-camera scenes with different geometry, occlusion patterns, and native frame rate.

\textbf{Transformer Engine mixed-precision fine-tuning:} we use NVIDIA Transformer Engine (TE) \cite{nvidia_transformer_engine_2025} to test whether mixed-precision fine-tuning can recover accuracy and stability beyond PTQ alone while still improving latency, thereby expanding the feasible camera-throughput region under reduced precision.

In addition, we introduce an identity-stability metric, \textbf{AvgTrackDur}, to complement standard detection and tracking scores by measuring the average duration of continuous identity tracks.

Overall, our goal is to provide actionable guidance for deploying Sparse4D-style architectures by quantifying trade-offs among accuracy, identity stability, latency, and training efficiency in realistic indoor MTMC settings.

\section{Related Work}
\label{sec::related}

\subsection{Spatiotemporal 3D Perception}
The transition from dense, map-centric representations to sparse, object-centric queries marks a pivotal shift in multi-view 3D detection. Dense approaches \cite{ huang2021bevdet, wang2022detr3d, li2024bevformer, teepe2024earlybird, Wang_2025_MCBLT} construct Bird's-Eye-View (BEV) grids, which scale quadratically with resolution and are often computationally prohibitive for real-time edge inference. 
In contrast, sparse query-based frameworks such as Sparse4D \cite{lin2022sparse4d, lin2023sparse4dv2, lin2023sparse4dv3} decouple structural awareness from feature density. Sparse4D utilizes 4D keypoints to iteratively refine object states by aggregating multi-view features across temporal windows. NVIDIA has re-engineered the Sparse4D architecture to adapt it for outside-in fixed infrastructure environments, as well as integrating object velocity estimations to improve tracking across occlusions \cite{wang2026unified3dobjectperception}; more details are provided in Section \ref{sec::baseline-model}.

\subsection{Multi-Camera Tracking \& Low-FPS Robustness}
Traditional MTMC tracking relies heavily on motion priors (e.g., Kalman Filters) to associate detections across frames. However, recent studies \cite{wang2024mcblt, zhou2022lowfps} indicate that motion-based association collapses in low-frame-rate (Low-FPS) regimes, where object displacement exceeds the search radius of standard filters.
While graph-based approaches like MCBLT \cite{wang2024mcblt} address this by modeling tracks as nodes in a global graph for long-term association, they often sacrifice end-to-end latency. Our research highlights the degradation of tracking accuracy at low FPS, necessitating more robust identification and velocity estimation techniques. 

\subsection{Quantization for Transformer Deployment}
Deploying Transformer-based detectors on edge accelerators such as an NVIDIA Orin often requires aggressive post-training quantization (PTQ). Compared to CNNs, Transformers can exhibit activation outliers that make INT8 quantization challenging \cite{yao2024comprehensive, xiao2023smoothquant}. Recent work on 3D perception models has also identified additional quantization failure modes tied to how geometric priors are injected into the network. For example, FQ-PETR \cite{yu2025fqpetr} reports a feature-scale imbalance during element-wise fusion, where geometry-related embeddings can dominate visual features and destabilize attention under low precision.

Sparse4D similarly integrates geometric information through world-coordinate reference points and multi-view projection/sampling. We therefore study PTQ and mixed-precision fine-tuning (ModelOpt \cite{nvidia_modelopt}, Transformer Engine \cite{nvidia_transformer_engine_2025}) to identify which parts of the pipeline are most sensitive to reduced precision and to quantify the resulting efficiency--accuracy trade-offs.

\section{Methodology}
\label{sec::methodology}

This section describes the evaluated Sparse4D-based baseline and the metrics used in our experiments.

\subsection{Baseline Model}
\label{sec::baseline-model}

\begin{figure*}[ht]
    \centering
    \includegraphics[width=\linewidth]{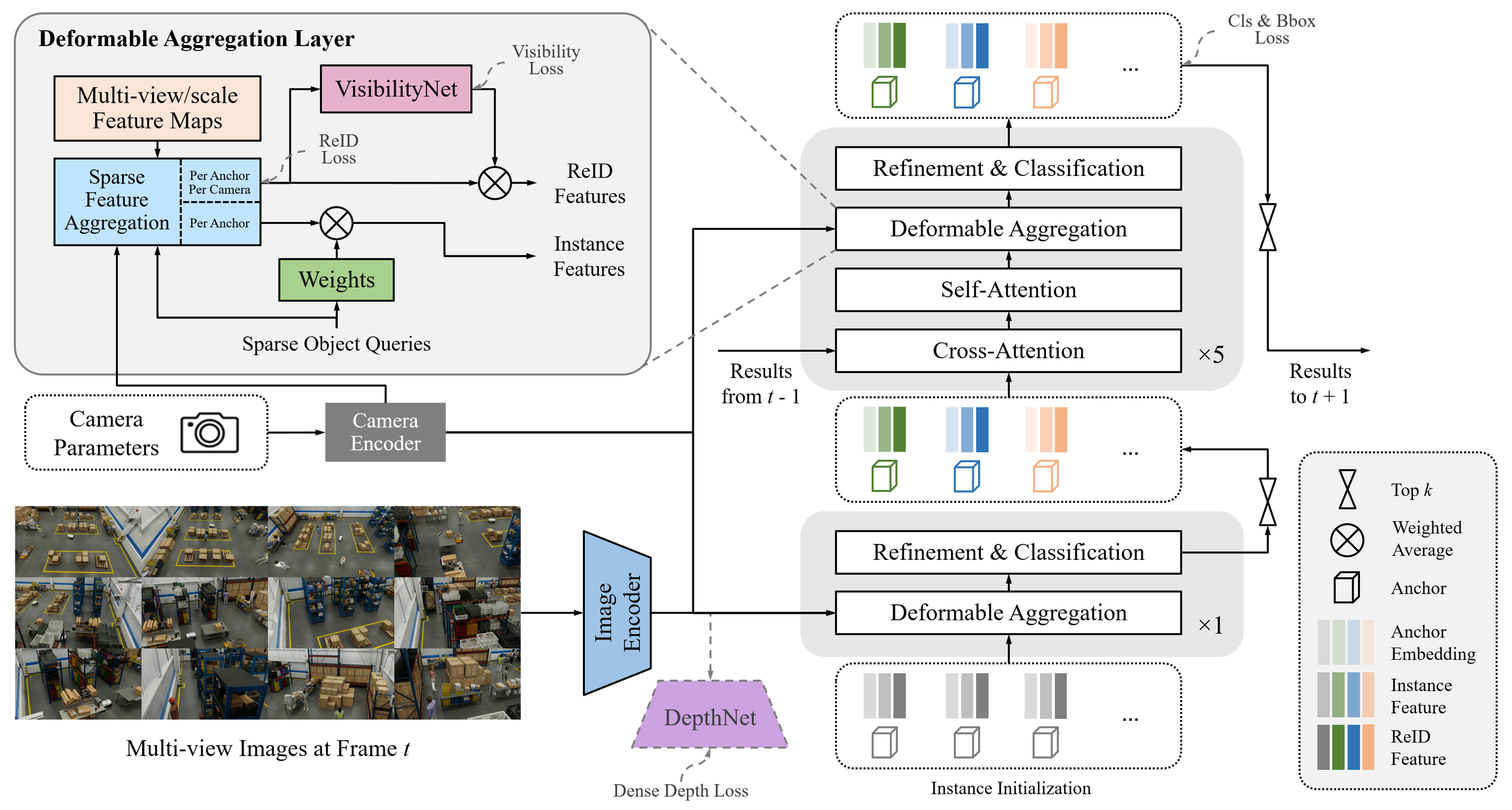}
    \caption{High-level overview of the Sparse4D-based model. Multi-view images and camera parameters are encoded into multi-scale feature maps. Sparse 3D queries aggregate multi-view evidence via deformable sampling and are updated by a spatiotemporal transformer with temporal memory. Prediction heads output per-object 3D state and an embedding used for association. This figure is originally from~\cite{wang2026unified3dobjectperception}.}
    \label{fig:architecture}
    \Description{Architecture diagram of the Sparse4D model}
\end{figure*}

Figure~\ref{fig:architecture} summarizes the Sparse4D-based unified 3D object perception model proposed by NVIDIA \cite{wang2026unified3dobjectperception}. At each timestep, the model consumes synchronized multi-view images with known camera intrinsics and extrinsics, producing multi-scale feature maps per camera via a shared image backbone/neck. The model represents objects using a set of sparse 3D queries expressed in a shared world coordinate frame. Each query carries a latent feature vector and is propagated across time using a temporal memory (instance bank), providing continuity through occlusions and viewpoint changes without relying on ego-motion.

To fuse information across cameras, each query is associated with 3D reference points (anchors) in world space. These points are projected into each view using camera calibration, and the model performs multi-view deformable sampling over the feature maps to extract view-conditioned evidence. A spatiotemporal transformer then updates the queries by combining current multi-view observations with the query memory. Finally, lightweight prediction heads regress 3D box state (e.g., center, size, yaw, velocity, confidence) and produce a per-object embedding, computed via visibility-weighted multi-view fusion, that supports temporal association and identity stability.

\subsection{Tracking Metrics and AvgTrackDur}

\paragraph{Standard Metrics.}
In addition to AvgTrackDur, we report standard metrics \cite{luiten2021hota}, including Higher Order Tracking Accuracy (HOTA), Detection Accuracy (DetA), Association Accuracy (AssA), and Localization Accuracy (LocA).



\paragraph{Definition.}
AvgTrackDur is computed from the same matching signals used by HOTA. For each timestep $t$, we perform Hungarian matching between predicted detections and ground-truth detections using an IoU threshold to establish frame-level correspondences. After matching, each predicted tracker ID $k$ is assigned a binary match indicator
\begin{equation}
M_k(t) = \mathbf{1}\{k \in \texttt{tracker\_ids}[t] \cap \texttt{gt\_ids}[t]\},
\end{equation}
which equals 1 if and only if tracker ID $k$ exists at time $t$ and is successfully matched to a ground-truth object. This definition intentionally ignores false positives by counting only timesteps where an identity is both present and correctly associated.

A \emph{run} for tracker ID $k$ is a maximal contiguous sequence of timesteps where $M_k(t)=1$. Let $D_{k,i} = t^{(k,i)}_e - t^{(k,i)}_s + 1$ denote the duration (in frames) of the $i$-th run for ID $k$, and let $\mathcal{R}$ denote the set of all runs across all tracker IDs. AvgTrackDur is then defined as:
\begin{equation}
\mathrm{AvgTrackDur} = \frac{\displaystyle\sum_{(k,i) \in \mathcal{R}} D_{k,i}}{|\mathcal{R}| \cdot F_0} \;\text{ seconds},
\label{eq:avgtrackdur}
\end{equation}
where $F_0$ is the reference frame rate used to convert frames to seconds.

\paragraph{Interpretation.}
Higher AvgTrackDur indicates longer continuous identity maintenance, while lower values indicate frequent identity breaks or switches. Because the metric is reported in seconds, it provides a direct and intuitive view of temporal stability that complements association-heavy metrics such as AssA.

\paragraph{Limitations.}
AvgTrackDur is dataset-dependent: longer sequences permit higher maximum durations, so comparisons across datasets should account for sequence length and scene dynamics. Tracks may also terminate naturally when objects exit the scene or become fully occluded, which penalizes high-turnover scenarios even under strong performance. Finally, AvgTrackDur does not directly assess localization quality; it measures only the duration of correct identity matches under the chosen IoU threshold.

\section{Model Optimization}
\label{sec::experiments}

This section describes the datasets and experimental protocols used to evaluate the model under the four deployment-oriented settings.

\subsection{Low-FPS Inference Robustness}
\label{subsec:method_lowfps}

We quantify how tracking quality and identity stability degrade when the effective inference frame rate is reduced, a common occurrence in bandwidth-limited deployments.

To simulate reduced inference frame rates, we skip frames in the data loader at a fixed stride. For example, retaining 1 out of every 5 frames reduces a 30 FPS stream to 6 FPS. We run inference across multiple inference rates and then evaluate tracking quality using a controlled evaluation setting. We only run this experiment on the AI City 2025 dataset due to its high native FPS of 30, whereas WILDTRACK is natively 2 FPS, which does not leave significant room to lower the FPS further. 

\paragraph{Controlled evaluation rate.}
The AI City evaluation framework can inflate tracking scores at lower sampling rates simply because fewer frames are scored. To ensure comparability of results across inference rates, we evaluate every run on the same fixed subset of frames at 1 FPS (1 out of every 30 frames), the only evaluation rate guaranteed to be present in all inference settings. For Warehouse 14, this corresponds to 300 evaluated frames out of 9000 total. We report HOTA, DetA, AssA, LocA, and AvgTrackDur to capture both tracking quality and long-horizon identity stability.

\subsection{Post-Training Quantization (PTQ)}
\label{subsec:method_ptq}

\begin{figure*}[t]
    \centering
    \includegraphics[width=0.85\textwidth]{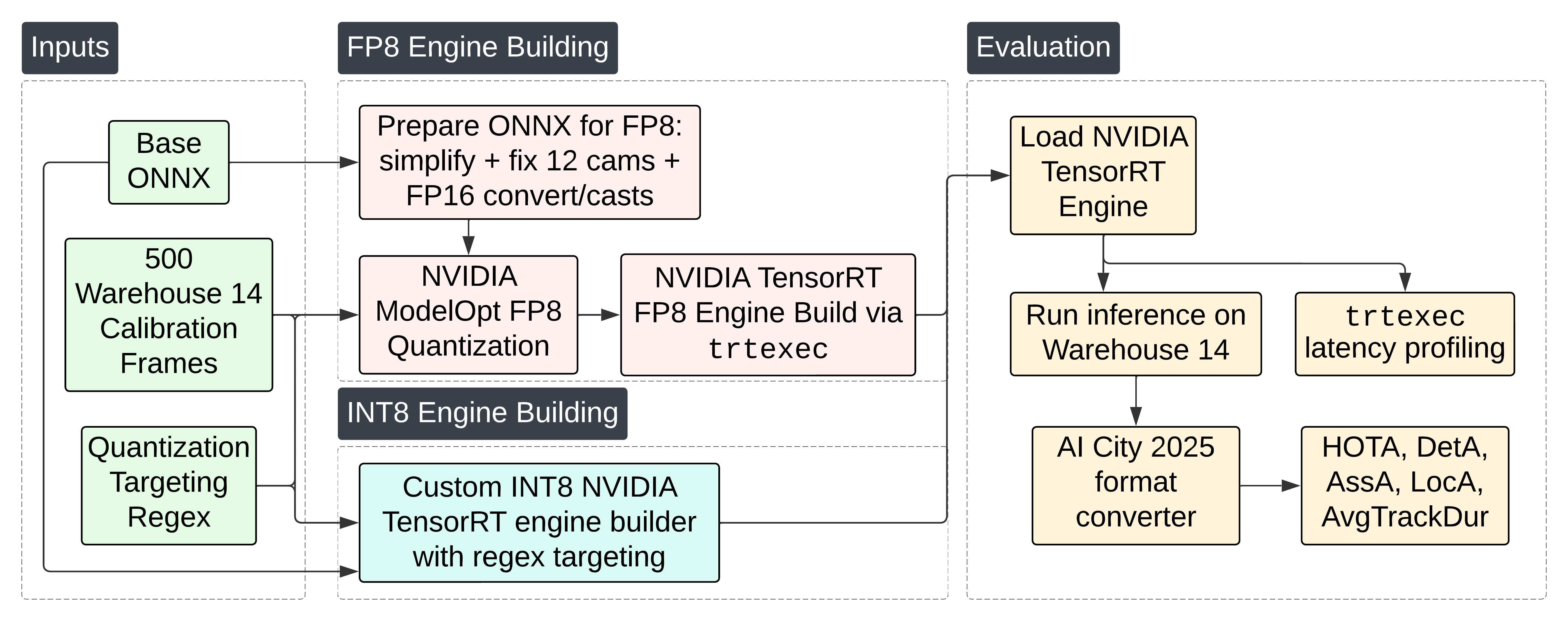}
    \caption{Overview of the INT8 and FP8 PTQ pipelines used to build TensorRT engines from the baseline ONNX and evaluate latency and tracking performance on AI City 2025 Warehouse 14.}
    \label{fig:ptq_flows}
    \footnotesize
    \Description{Flow diagram with inputs on the far left that include the ONNX file, the calibration data, and the regex of the layers to quantize. In the middle, there are the FP8 and INT8 engine building blocks. FP8 utilizes ModelOpt, while INT8 employs a custom script. On the far right is the evaluation section where the engines are loaded and then run on Warehouse 14 to be evaluated using the AI City 2025 evaluation pipeline.}
\end{figure*}

We evaluate the effects of post-training quantization on both tracking performance and latency across two quantization methodologies to identify the optimal trade-off between speed and accuracy, as well as which submodules are most sensitive to quantization.

\paragraph{INT8 PTQ (A100).}
We build TensorRT engines directly from the baseline ONNX using a custom engine builder with regex-based layer targeting. Activation scales are calibrated using the first 500 frames from the AI City 2025 Warehouse 14 training split. We sweep quantization scope by modifying the target regex (for example, backbone-only, backbone+neck, and backbone+neck+attention), while leaving non-targeted layers in higher precision (FP16). Each INT8 engine is benchmarked for frame rate (FPS), then evaluated end-to-end on Warehouse 14 by converting outputs to the AI City competition format and computing HOTA, DetA, AssA, LocA, and AvgTrackDur.

\paragraph{FP8 PTQ (H100/H200-class).}
For FP8 PTQ, we first prepare an FP8-friendly ONNX by (i) simplifying the graph with \texttt{onnxsim}, (ii) fixing the camera dimension to a 12-camera static shape, and (iii) converting to FP16 while inserting targeted cast nodes for numerically sensitive operators. We then quantize the prepared ONNX using NVIDIA ModelOpt \cite{nvidia_modelopt} with 500 Warehouse 14 calibration frames, restricting quantization to selected operator types (for example Conv/MatMul/Gemm) and optionally to submodules via regex. The resulting FP8-annotated ONNX is compiled into a strongly-typed TensorRT engine using \texttt{trtexec} \cite{nvidia_tensorrt, nvidia_trtexec}. As with INT8, we evaluate each FP8 variant using the same downstream tracking pipeline and report inference latency and tracking metrics.

\paragraph{Inference latency measurement and hardware comparability.}
We report INT8 latency on NVIDIA A100 (Ampere) and FP8 latency on H100/H200-class GPUs (Hopper), matching each platform’s native Tensor Core support. Tracking accuracy metrics (HOTA, DetA, AssA, LocA, AvgTrackDur) should be largely hardware-invariant, but FPS should only be compared within the same GPU family. Figure~\ref{fig:ptq_flows} summarizes both PTQ workflows.

\subsection{WILDTRACK Adaptation and Fine-Tuning}
\label{subsec:method_wildtrack}

We evaluate Sparse4D’s generalization to the WILDTRACK dataset and assess whether fine-tuning improves detection and tracking performance under WILDTRACK’s low temporal resolution (2 FPS). A central contribution of this work is the development of a complete adaptation pipeline that enables the Sparse4D-based baseline to operate on WILDTRACK without architectural modifications.

\paragraph{Format Adaptation}
WILDTRACK differs substantially from the AI City warehouse domain in camera placement, spatial extent, annotation format, and scene statistics. To bridge this gap, we implement a custom dataset conversion pipeline that transforms the original grid-based positionID annotations into a TAO-compatible format. Each positionID is mapped to metric $(x, y)$ coordinates, the region of interest is recentered, and 3D bounding boxes are constructed with centers placed at half the assumed person height above the ground plane. For each frame, our pipeline tracks person identities over time to estimate per-object velocities and preserves the original 2D bounding boxes for all visible cameras.


\paragraph{Anchor Bank Regeneration}
In addition to annotation conversion, we regenerate the 3D anchor bank used by Sparse4D to better reflect the spatial statistics of WILDTRACK. The default anchor configuration, derived from AI City warehouse scenes, provides suboptimal coverage due to differences in scale, region of interest, and pedestrian density. We therefore initialize a new anchor bank directly from WILDTRACK ground-truth boxes using $k$-means clustering over the $(x, y, z)$ centers of all training-set annotations, with $K=900$ anchors to match the default Sparse4D configuration. 

\paragraph{Training Modifications}
We evaluate a sequence of fine-tuning strategies to isolate which adaptations are most impactful for WILDTRACK. Since the dataset contains only axis-aligned pedestrians, we rebalance regression losses by disabling yaw and box-dimension terms and focusing optimization on center coordinates. To stabilize training on the small dataset, we use reduced learning rates (e.g., $10^{-5}$) to limit catastrophic forgetting from pretrained warehouse weights. We also evaluate a person-only classification head to reduce gradient noise from unused classes, and apply region-of-interest filtering and confidence thresholding during post-processing to suppress spurious detections and improve assignment stability under low-FPS conditions.

\begin{figure}[htbp]
\centering
\includegraphics[width=\linewidth]{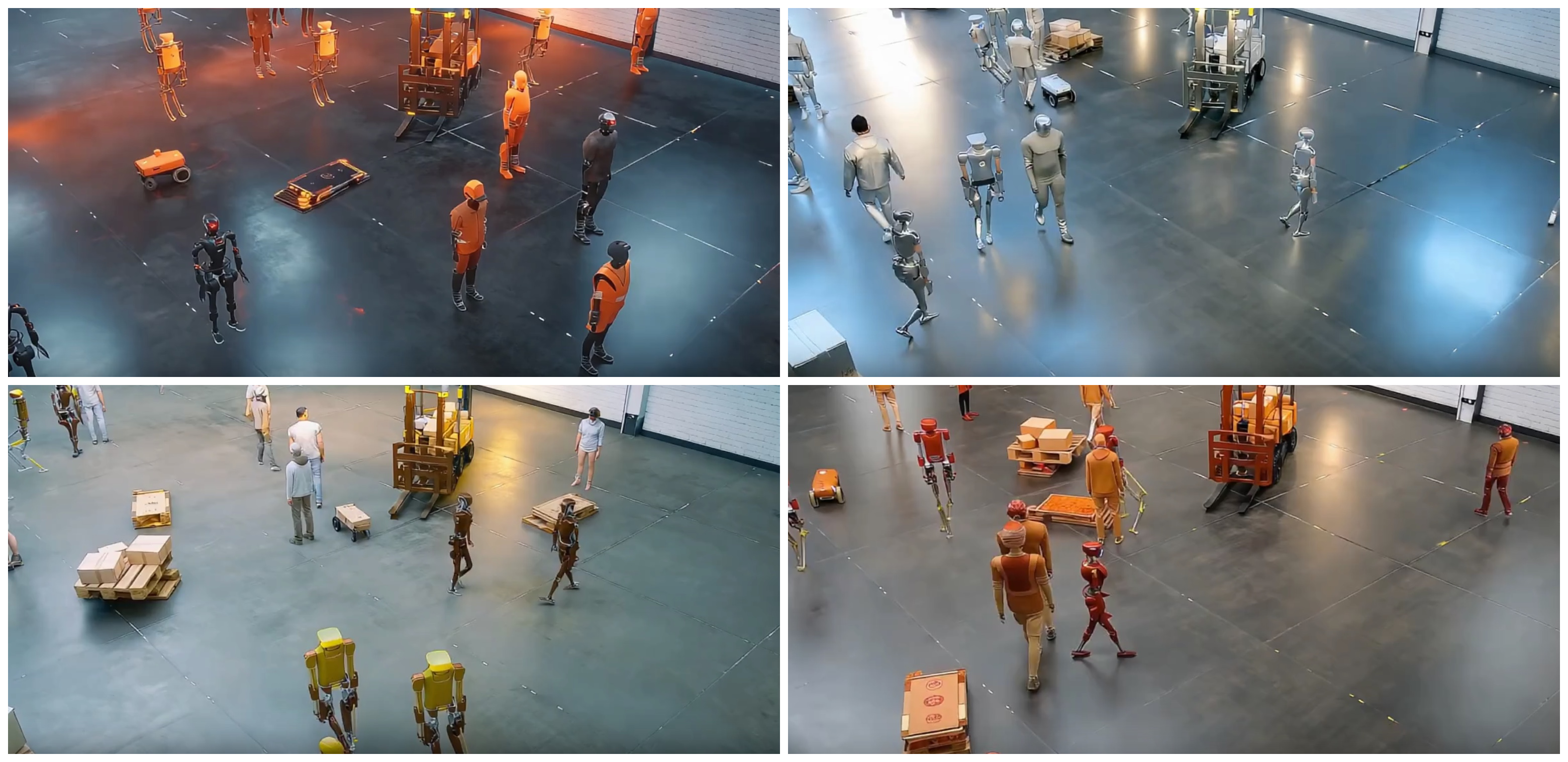}
\caption{Examples of appearance styles generated using COSMOS Transfer 1 for synthetic MTMC training clips.}
\label{fig:cosmos_styles}
\Description{Four images of similar scenes generated using different coloring and lighting styles.}
\end{figure}

\paragraph{COSMOS Checkpoint}
Finally, to better match WILDTRACK’s temporal regime, we evaluate a COSMOS-style low-FPS pretrained checkpoint trained on synthetic data augmented with COSMOS Transfer 1 appearance variations and a multi-FPS curriculum spanning 1–30 FPS. As illustrated in Figure~\ref{fig:cosmos_styles}, single-camera clips are segmented into multiple prompt-stylized variants that are recombined into full sequences while reusing the original synthetic ground truth. We evaluate both zero-shot transfer from this checkpoint and further WILDTRACK-specific fine-tuning initialized from it. 

\subsection{Transformer Engine Mixed-Precision Fine-Tuning}
\label{subsec:method_te}

We evaluate the ability of Transformer Engine (TE) \cite{nvidia_transformer_engine_2025} to improve training and inference efficiency using mixed precision on Sparse4D. This experiment complements PTQ by allowing the model to adapt under the low-precision environment, rather than simply applying quantization after the fact.

\paragraph{Integration and precision targeting.}
We integrate NVIDIA Transformer Engine into the training pipeline to enable mixed-precision execution where selected operations run in FP8 while weights are stored in FP16. To balance stability and speed, we target the first single-frame decoder layer and the anchor encoder for FP8 compute while leaving the remaining layers in standard precision.

\paragraph{Training setup.}
We fine-tune the model for 22{,}500 iterations using batch size 6 across 2 GPUs at $960 \times 540$ resolution. We use TE's hybrid FP8 mode (\texttt{HYBRID}) with an \texttt{amax} history length of 16 for dynamic scaling.

\paragraph{Export and benchmarking.}
After fine-tuning, we export ONNX and benchmark inference latency using \texttt{trtexec} \cite{nvidia_trtexec} on an NVIDIA H200. We report mean latency and FPS for a standard 12-camera configuration, and we conduct a camera scalability stress test by increasing the number of camera streams until the mean inference latency is too high for real-time operation (30 FPS). In addition to latency, we perform qualitative inspection of consecutive frames to detect potential identity propagation failures that may not be captured by detection confidence alone.

\section{Experimental Results}
\label{sec::results}


\subsection{Datasets}

\paragraph{AI City Challenge 2025 Dataset}
The AI City Challenge 2025 dataset consists of multi-camera video sequences set in warehouse and logistics environments, annotated for multi-target multi-camera (MTMC) tracking and 3D perception. The scenes are recorded by multiple static cameras with overlapping or partially overlapping fields of view and include ground-truth identities, cross-camera correspondences, and 3D bounding box annotations. As a synthetic benchmark, AI City provides clean, consistent labels and controlled capture conditions, making it well-suited for isolating model behavior under systematic changes such as reduced effective frame rate and post-training quantization.

Building on the multi-camera warehouse settings introduced in the 2024 edition, the 2025 dataset expands the set of tracked object categories beyond pedestrians. Annotations now include several classes of service robots and mobile platforms: forklifts, NovaCarter robots, Transporter robots, FourierGR1T2 humanoid robots, and AgilityDigit humanoid robots. This diversity provides a challenging benchmark for assessing robustness across object types with varying sizes, motion patterns, and appearance characteristics.

\paragraph{WILDTRACK Dataset}
The WILDTRACK dataset contains synchronized video recordings from seven static cameras observing a shared outdoor pedestrian walkway. All cameras are precisely calibrated, and the dataset provides ground-plane annotations for pedestrian locations, enabling accurate multi-view geometric reasoning. In contrast to AI City, WILDTRACK is captured in the real world and therefore includes practical imaging and system effects (e.g., lighting changes, motion blur, compression artifacts, occlusions, and minor calibration or synchronization imperfections) that can degrade geometric projection and temporal identity maintenance.

Unlike the warehouse sequences in AI City, WILDTRACK is captured at 2 FPS and contains only pedestrians in a compact spatial region. These properties make it a useful stress test for temporal reasoning under low frame rates and for evaluating domain transfer from warehouse to outdoor settings, which is closer to real deployment conditions.

For WILDTRACK, we fine-tune on the first 360 frames and test on the remaining 40 frames.

\subsection{Low-FPS Inference Robustness}
\label{subsec:results_lowfps}

Table~\ref{tab:fps_tracking_metrics} shows stable tracking from 30 FPS down to 3 FPS (HOTA near 43\% and high AvgTrackDur), followed by a sharp degradation below 2 FPS. The drop is primarily association-driven: AssA decreases from 43.6\% (3 FPS) to 36.5\% (2 FPS) and 27.8\% (1 FPS), while DetA remains comparatively stable at 43.0\% (3 FPS), 43.1\% (2 FPS), and 42.4\% (1 FPS). AvgTrackDur highlights the break in identity persistence, falling from 268.5 seconds (3 FPS) to 199.3 seconds (2 FPS) and 156.6 seconds (1 FPS). This indicates detection remains largely intact while temporal continuity fails as inter-frame motion grows relative to the model's propagation and association mechanisms.

\begin{table}[htbp]
    \centering
    \caption{Warehouse 14 (AI City 2025): Tracking metrics across inference frame rates.}
    \label{tab:fps_tracking_metrics}
    \small
    \begin{tabular}{cccccc}
        \toprule
        Inference FPS & HOTA & DetA & AssA & LocA & AvgTrackDur (s) \\
        \midrule
        30 & 42.9 & 42.7 & 43.1 & 51.9 & 251.6 \\
        15 & 42.9 & 42.7 & 43.1 & 47.5 & 298.5 \\
        10 & 43.0 & 42.8 & 43.2 & 50.4 & 298.5 \\
         6 & 43.0 & 42.8 & 43.3 & 55.7 & 298.5 \\
         5 & 43.1 & 42.9 & 43.3 & 50.7 & 298.5 \\
         3 & 43.3 & 43.0 & 43.6 & 47.4 & 268.5 \\
         2 & 39.4 & 43.1 & 36.5 & 53.4 & 199.3 \\
         1 & 31.1 & 42.4 & 27.8 & 58.4 & 156.6 \\
        \bottomrule
    \end{tabular}
\end{table}

When the results are analyzed per class, the more mobile classes, such as Person and Transporter, show a decrease in AssA and AvgTrackDur, while the static classes do not.

\subsection{Post-Training Quantization (PTQ)}
\label{subsec:results_ptq}

\paragraph{PTQ Reporting Conventions}
We report class-averaged tracking metrics (in \%) on Warehouse 14; AvgTrackDur is in seconds for the \textit{Person} class only. BB, NK, and AT denote the Backbone, Neck, and Attention sections of the model, respectively. We evaluate using the first 3000 frames of Warehouse 14 (100 seconds).

\paragraph{INT8 PTQ (A100)}
Table~\ref{tab:quantization_metrics} reports INT8 results on Warehouse 14. Early backbone quantization yields modest latency improvements with moderate accuracy loss, while more aggressive quantization increases FPS but substantially degrades tracking.

Quantizing Backbone+Neck increases inference frame rate from 28.2 to 36.1 FPS and improves HOTA from 47.0\% to 47.8\%, but AvgTrackDur collapses from 60.0 seconds to 12.3 seconds (Person class), indicating mild fragmentation despite strong aggregate scores. Extending quantization into attention degrades both aggregate tracking and stability (HOTA 35.2\%, AvgTrackDur 5.7 seconds), suggesting attention layers are more sensitive than the backbone and neck layers. 

\begin{table}[htbp]
    \centering
    \caption{Class-averaged tracking metrics (in \%) for INT8 quantization configurations. INT8 Cnt denotes the number of INT8 layers (out of 1394 total).}
    \label{tab:quantization_metrics}
    \small
    \setlength{\tabcolsep}{1.75pt}
    \begin{tabular}{@{}>{\raggedright\arraybackslash}p{0.21\columnwidth}ccccccc@{}}
        \toprule
        Quant. Layers        & INT8 Cnt  & FPS & HOTA & DetA & AssA & LocA & AvgTrackDur \\
        \midrule
        Baseline                & 0                 & 28.2 & 47.0 & 46.5 & 47.7 & 48.2 & 60.0 \\
        \addlinespace[2pt]
        BB Layer 1        & 51                & 28.8 & 44.6 & 44.3 & 45.0 & 55.9 & 75.0 \\
        \addlinespace[2pt]
        BB Layers 1--3    & 515               & 33.1 & 44.4 & 43.9 & 45.0 & 58.5 & 42.7 \\
        \addlinespace[2pt]
        BB Layers 1--4    & 571               & 33.9 & 45.7 & 45.2 & 46.2 & 59.3 & 74.8 \\
        \addlinespace[2pt]
        BB + NK        & 580               & 36.1 & 47.8 & 47.0 & 48.9 & 57.9 & 12.3 \\
        \addlinespace[2pt]
        BB + NK + AT  & 670               & 31.4 & 35.2 & 35.2 & 36.7 & 54.0 & 5.7 \\
        \bottomrule
    \end{tabular}
\end{table}

\paragraph{FP8 PTQ (H100)}
Table~\ref{tab:fp8-quantization_metrics} reports FP8 PTQ results on Hopper-class hardware. Quantizing Backbone+Neck improves frame rate from 68.9 to 77.8 FPS with essentially unchanged aggregate tracking. Extending FP8 into attention preserves HOTA but reduces localization accuracy (LocA 56.7\% to 55.3\%), again indicating higher numerical sensitivity in attention-related components. AvgTrackDur remains mostly constant across FP8 settings, correlating well with the lack of any drop in AssA.

\paragraph{Note on PTQ baselines}
The INT8 and FP8 “baseline” rows are generated from different pipelines, so their absolute scores are not directly comparable across tables. The FP8 pipeline applies an ONNX simplification step for ModelOpt compatibility, which rewrites the graph, reducing the effective layer count (e.g., 836 vs.\ 1394), and also produces the FP16 reference via ModelOpt. In contrast, the INT8 baseline is evaluated on the original graph using engine-level FP16 settings without ModelOpt. These pipeline differences cause baseline accuracy shifts independent of FP8 vs.\ INT8 quantization.


\begin{table}[htbp]
    \centering
    \caption{Class-averaged tracking metrics (in \%) for FP8 quantization configurations. Layers were targeted using regex. FP8 Cnt is the number of FP8 layers (out of 836 total).}
    \label{tab:fp8-quantization_metrics}
    \small 
    \setlength{\tabcolsep}{1.75pt}
    \begin{tabular}{@{}>{\raggedright\arraybackslash}p{0.225\columnwidth}ccccccc@{}}
        \toprule
        Quant. Layers           & FP8 Cnt   & FPS   & HOTA & DetA & AssA & LocA & AvgTrackDur \\
        \midrule
        Baseline                & 0         & 68.9  & 44.2 & 43.9 & 44.6 & 58.6 & 59.9 \\
        \addlinespace[2pt]
        BB + NK                 & 102       & 77.8  & 44.2 & 43.9 & 44.7 & 56.7 & 49.9 \\
        \addlinespace[2pt]
        BB + NK + AT            & 142       & 77.3  & 44.3 & 43.9 & 44.7 & 55.3 & 59.96 \\
        \bottomrule
    \end{tabular}
\end{table}

\subsection{WILDTRACK Adaptation and Fine-Tuning}
\label{subsec:results_wildtrack}

\begin{table*}[ht]
\centering
\caption{Person-class AP on the WILDTRACK test split.}
\label{tab:ap_results}
\small
\resizebox{0.85\linewidth}{!}{
\begin{tabular}{l c c p{7.0cm}}
\hline
\textbf{Model Checkpoint} 
& \textbf{AP (Person)} 
& \textbf{LR} 
& \textbf{Fine-Tuning / Configuration Details} \\
\hline
Sparse4D Base (Zero-shot) 
& 0.751 
& $1\times10^{-3}$ 
& Default checkpoint trained on AI City warehouse scenes; no WILDTRACK-specific adaptation. \\

Sparse4D Fine-tuned 
& 0.751 
& $1\times10^{-5}$ 
& Loss rebalancing: regression weights for yaw and box dimensions set to zero; optimized center coordinates. \\

Sparse4D + WILDTRACK Anchors 
& 0.751 
& $8.5\times10^{-4}$ 
& Anchor bank regenerated using k-means clustering over WILDTRACK boxes. Ran on base checkpoint. \\

Sparse4D (Person-only Weights) 
& 0.751 
& $1\times10^{-5}$ 
& Classification heads pruned to retain only the person class. \\

COSMOS Low-FPS Checkpoint 
& 0.856 
& $1\times10^{-5}$ 
& Pretrained with COSMOS-style appearance augmentation and multi-FPS curriculum (1--30 FPS); evaluated zero-shot. \\

COSMOS + Fine-tuning 
& 0.857 
& $1\times10^{-5}$ 
& Initialized from COSMOS checkpoint; WILDTRACK-specific fine-tuning and post-processing. \\
\hline
\end{tabular}
}
\end{table*}

\paragraph{Detection performance (AP)}
Table~\ref{tab:ap_results} compares initialization and fine-tuning strategies. The base checkpoint achieves an AP of 0.751. Several WILDTRACK-specific fine-tuning attempts (loss rebalancing, WILDTRACK-specific anchors, and person-only classifier pruning) do not improve AP over the base model. In contrast, a COSMOS-style low-FPS checkpoint generalizes strongly zero-shot (AP = 0.856), and additional fine-tuning yields a small gain (AP = 0.857). This suggests that pretraining for appearance diversity and variable frame rates is more impactful than small-scale in-domain fine-tuning for WILDTRACK under the current setup and data volume.

\noindent\textit{Note:} Although COSMOS-style pretraining improves AP, prior work achieves higher 
WILDTRACK performance, suggesting that additional architectural or training refinements beyond our current setup are necessary.


\begin{figure*}[htbp]
\centering
\includegraphics[width=\linewidth]{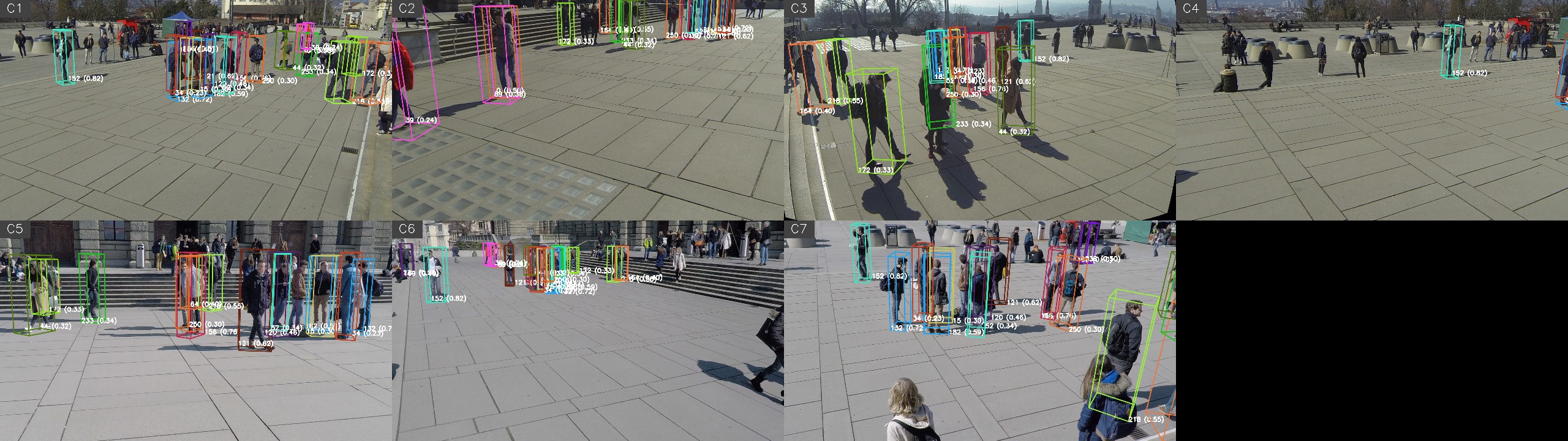}
\caption{Zero-shot detection visualization on WILDTRACK test frame 360 (confidence threshold 0.2), using the COSMOS low-FPS checkpoint.}
\label{fig:cosmos_wildtrack_vis}
\Description{WILDTRACK visualization with colored boxes encapsulating each person. Next to each box is a white number representing the confidence of the box from 0 to 100\%.}
\end{figure*}

\paragraph{Tracking behavior under 2 FPS}
Tracking remains challenging on WILDTRACK at 2 FPS. This aligns with Section~\ref{subsec:results_lowfps}, where association degrades sharply below 2 FPS even when detection quality remains strong. By applying tracking-focused post-processing in our WILDTRACK pipeline, we improve HOTA from 3.9 to 6.3, indicating that assignment heuristics such as ROI filtering and confidence thresholding can help preserve identity continuity at low FPS.

\paragraph{Impact of anchor bank design}
This regenerated anchor bank improves spatial coverage and recall consistency but does not fully resolve tracking failures induced by the dataset’s low frame rate, indicating that anchor alignment alone is insufficient without stronger temporal cues. 

\subsection{Transformer Engine Mixed-Precision Fine-Tuning}
\label{subsec:results_te}

\paragraph{Inference Latency and Frame Rate (12 Cameras)}
Table~\ref{tab:fp8_results} compares baseline FP16 inference against TE mixed precision on an NVIDIA H200 with a standard 12-camera input. TE reduces mean latency from 24.24 ms to 16.39 ms and increases inference frame rate from 41.25 FPS to 61.02 FPS, achieving a 48\% speedup.

\begin{table}[htbp]
\centering
\caption{Comparison of mean inference latency and frame rate on NVIDIA H200 (12-camera input). FP16/FP8 refers to storing the weights in FP16 and then converting to FP8 for the computation.}
\label{tab:fp8_results}
\small
\setlength{\tabcolsep}{4pt}
\begin{tabular}{lcccc}
\toprule
\textbf{Config} & \textbf{Prec.} & \textbf{Mean Lat. (ms)} & \textbf{FPS} & \textbf{Speedup} \\
\midrule
Baseline (R101) & FP16 & 24.24 ms & 41.25 FPS & - \\
TE & FP16/FP8 & \textbf{16.39 ms} & \textbf{61.02 FPS} & \textbf{+48\%} \\
\bottomrule
\end{tabular}
\end{table}

\paragraph{Camera Scalability Stress Test}
Table~\ref{tab:cam_scaling} demonstrates the increased camera capacity per GPU. The baseline configuration fails to meet real-time requirements ($>30$ FPS) at 18 cameras. In contrast, the TE Mixed-Precision model remains above 30 FPS up to 30 cameras (31.40 FPS). This represents a 66\% increase in camera capacity (from 18 to 30) under the same real-time constraint.

\begin{table}[htbp]
\centering
\caption{Transformer Engine scalability stress test: inference performance vs.\ camera count on a single NVIDIA H200. The RT column indicates real-time performance (> 30 FPS).}
\label{tab:cam_scaling}
\small
\setlength{\tabcolsep}{4pt}
\begin{tabular}{@{}lcccc@{}}
\toprule
\textbf{Config} & \textbf{Cams} & \textbf{Latency (ms)} & \textbf{FPS} & \textbf{RT} \\
\midrule
Baseline (FP16) & 18 & 34.16 & 29.27 & \textcolor{red}{\ding{55}} \\
\midrule
\multirow{5}{*}{FP8 (MP)} 
 & 18 & \textbf{21.40} & \textbf{46.74} & \textcolor{green}{\ding{51}} \\
 & 21 & 22.17 & 45.11 & \textcolor{green}{\ding{51}} \\
 & 24 & 26.22 & 38.14 & \textcolor{green}{\ding{51}} \\
 & 27 & 29.05 & 34.42 & \textcolor{green}{\ding{51}} \\
 & 30 & 31.85 & 31.40 & \textcolor{green}{\ding{51}} \\
\bottomrule
\end{tabular}
\end{table}

\begin{figure}[htbp]
    \centering
    \begin{subfigure}[b]{\linewidth}
        \includegraphics[width=\linewidth]{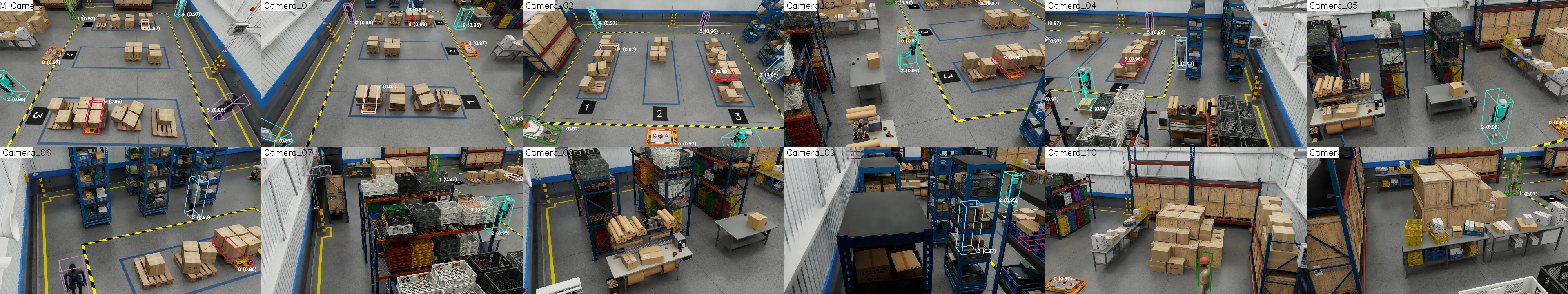}
        \caption{Frame $t$: High-confidence detections (bbox confidence $>0.95$) with initial ID assignments.}
        \label{fig:fp8_frame0}
        \Description{Frame $t$ detections on the Warehouse 14 dataset, colored boxes enclose each robot and person in the scene.}
    \end{subfigure}
    \vspace{0.2cm}
    \begin{subfigure}[b]{\linewidth}
        \includegraphics[width=\linewidth]{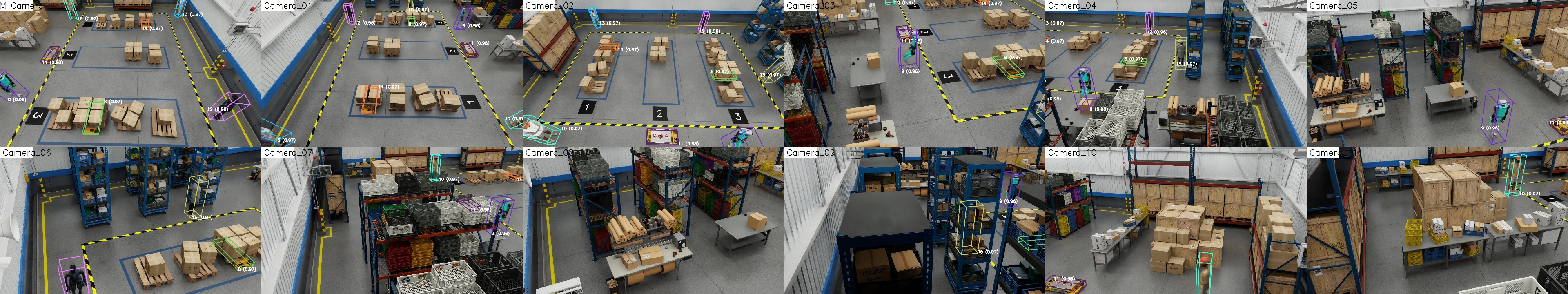}
        \caption{Frame $t+1$: Robust detection persists with accurate bounding boxes, but IDs (indicated by box colors) switch immediately, replacing the previous frame's identities.}
        \label{fig:fp8_frame1}
        \Description{Frame $t+1$ detections on the Warehouse 14 dataset, colored boxes enclose each robot and person in the scene. These colored boxes are different than the colors in frame $t$.}
    \end{subfigure}
    \caption{Qualitative visualization of the Mixed-Precision model on the NVIDIA H200.}
    \label{fig:fp8_qualitative}
\end{figure}

Despite the large latency reduction, detections remain accurate, but identity propagation breaks across frames (Figure~\ref{fig:fp8_qualitative}), with new IDs assigned each timestep even for stationary objects. This suggests a mismatch between cached instance-memory queries and new-frame features under mixed-precision execution, consistent with our broader finding that acceleration can preserve detection quality while degrading identity stability.



\section{Discussion}
\label{sec::discussion}

\subsection{Limitations}

\textbf{Hardware heterogeneity.} INT8 latency is reported on A100, while FP8 and Transformer Engine results are reported on H100/H200-class GPUs, causing FPS values to be non-comparable across hardware. 

\textbf{PTQ cross-precision comparability.} The INT8 and FP8 PTQ results use different pipelines (graph preparation and FP16 reference generation), so they should not be interpreted as a direct INT8-versus-FP8 comparison. A more controlled study would run both INT8 and FP8 through a single standardized ModelOpt workflow so that baselines and accuracy deltas are comparable. Our goal is to evaluate the effects of different PTQ approaches rather than to claim INT8 is better or worse than FP8.

\textbf{Limited WILDTRACK fine-tuning data.} WILDTRACK provides only 360 training frames, and its object density and motion patterns differ greatly from the warehouse scene. This limits the amount of task-specific supervision available for fine-tuning and can encourage overfitting, which may help explain the limited gains we observe. We also emphasize detection AP on WILDTRACK; integrating HOTA, AssA, and AvgTrackDur on WILDTRACK is needed for stronger conclusions about identity stability in that domain.

\subsection{Future Work}
Based on the observed trends, the next steps that appear most likely to improve practical deployment readiness are:

\textbf{Attention sensitivity profiling.} Perform a systematic layer-wise sweep where attention sub-blocks are quantized or FP8-enabled incrementally, measuring not only HOTA but also AvgTrackDur and short-window ID survival.

\textbf{Low-FPS association training.} Evaluate fine-tuning recipes that explicitly target low-FPS behavior, including frame-gap sampling, longer temporal unroll, and association-supervised objectives where possible.

\textbf{Instance bank debugging under mixed precision.} Add instrumentation to log matching scores between cached queries and new detections, and compare FP16 versus FP8 compute paths to localize where the representation shift causes cached tracks to be rejected.

\textbf{Unified deployment metric suite.} Standardize a compact set of deployment-facing metrics: FPS (per-camera effective), HOTA/DetA/AssA, and AvgTrackDur, reported together across all acceleration variants.

\section{Conclusion}
\label{sec::conclusion}

This paper evaluated Sparse4D as a deployment-oriented baseline for outside-in indoor multi-camera 3D perception, emphasizing four practical experiment tracks: low-FPS inference, post-training quantization, WILDTRACK adaptation, and Transformer Engine mixed-precision fine-tuning. Across experiments, the central limiting factor was not raw detection quality but temporal association stability, which we made easier to interpret with AvgTrackDur.

We find that Sparse4D remains comparatively robust down to moderate inference frame rates, but tracking degrades sharply at very low FPS, driven primarily by drops in association accuracy and track fragmentation. For acceleration, selective PTQ (backbone and neck) provided the best observed latency-accuracy trade-off, while applying low precision to attention layers consistently behaved as a sensitivity hotspot. FP8 PTQ better preserved localization and association accuracy than INT8 PTQ when targeting attention layers. On WILDTRACK, conventional fine-tuning did not yield consistent AP gains over the base checkpoint; in contrast, low-FPS pretraining (COSMOS-style) improved zero-shot performance, suggesting that temporal regime alignment matters more than shallow dataset-specific adaptation. Finally, Transformer Engine mixed precision delivered large latency and camera-scaling gains on H200-class hardware, but surfaced identity propagation instability that must be resolved before deployment.

Sparse4D is a strong baseline for indoor MTMC deployments, and substantial speedups are achievable without immediate loss in aggregate tracking quality. However, the primary failure mode under low FPS and aggressive acceleration is association instability rather than detection quality. Future work should focus on association-aware training and precision-aware stability analysis to preserve identity continuity while maintaining these throughput gains.

\section*{Acknowledgments}
This work was completed as part of the Clemson University School of Computing Senior Capstone Program. We thank NVIDIA for their support and collaboration throughout the semester. We especially thank Prof.~Russell, Roopa, and Karthick for orchestrating the NVIDIA--Clemson collaboration that made this project possible.

\FloatBarrier

\appendix

\setcitestyle{numbers}
\bibliographystyle{plainnat}
\bibliography{main}

@InProceedings{Wang_2025_MCBLT,
    author    = {Wang, Yizhou and Meinhardt, Tim and Cetintas, Orcun and Yang, Cheng-Yen and Pusegaonkar, Sameer and Missaoui, Benjamin and Biswas, Sujit and Tang, Zheng and Leal-Taixe, Laura},
    title     = {MCBLT: Multi-Camera Multi-Object 3D Tracking in Long Videos},
    booktitle = {Proceedings of the IEEE/CVF International Conference on Computer Vision (ICCV) Workshops},
    month     = {October},
    year      = {2025},
    pages     = {5245-5254}
}

@article{lin2022sparse4d,
  title={Sparse4d: Multi-view 3d object detection with sparse spatial-temporal fusion},
  author={Lin, Xuewu and Lin, Tianwei and Pei, Zixiang and Huang, Lichao and Su, Zhizhong},
  journal={arXiv preprint arXiv:2211.10581},
  year={2022}
}

@article{lin2023sparse4dv2,
  title={Sparse4d v2: Recurrent temporal fusion with sparse model},
  author={Lin, Xuewu and Lin, Tianwei and Pei, Zixiang and Huang, Lichao and Su, Zhizhong},
  journal={arXiv preprint arXiv:2305.14018},
  year={2023}
}

@article{lin2023sparse4dv3,
  title={Sparse4d v3: Advancing end-to-end 3d detection and tracking},
  author={Lin, Xuewu and Pei, Zixiang and Lin, Tianwei and Huang, Lichao and Su, Zhizhong},
  journal={arXiv preprint arXiv:2311.11722},
  year={2023}
}

@inproceedings{teepe2024earlybird,
  title={EarlyBird: early-fusion for multi-view tracking in the bird's eye View},
  author={Teepe, Torben and Wolters, Philipp and Gilg, Johannes and Herzog, Fabian and Rigoll, Gerhard},
  booktitle={Proceedings of the IEEE/CVF Winter Conference on Applications of Computer Vision},
  pages={102--111},
  year={2024}
}

@inproceedings{zheng2019joint,
  title={Joint discriminative and generative learning for person re-identification},
  author={Zheng, Zhedong and Yang, Xiaodong and Yu, Zhiding and Zheng, Liang and Yang, Yi and Kautz, Jan},
  booktitle={proceedings of the IEEE/CVF conference on computer vision and pattern recognition},
  pages={2138--2147},
  year={2019}
}

@inproceedings{wang2022detr3d,
  title={Detr3d: 3d object detection from multi-view images via 3d-to-2d queries},
  author={Wang, Yue and Guizilini, Vitor Campagnolo and Zhang, Tianyuan and Wang, Yilun and Zhao, Hang and Solomon, Justin},
  booktitle={Conference on robot learning},
  pages={180--191},
  year={2022},
  organization={PMLR}
}

@article{huang2021bevdet,
  title={Bevdet: High-performance multi-camera 3d object detection in bird-eye-view},
  author={Huang, Junjie and Huang, Guan and Zhu, Zheng and Ye, Yun and Du, Dalong},
  journal={arXiv preprint arXiv:2112.11790},
  year={2021}
}

@article{li2024bevformer,
  title={Bevformer: learning bird's-eye-view representation from lidar-camera via spatiotemporal transformers},
  author={Li, Zhiqi and Wang, Wenhai and Li, Hongyang and Xie, Enze and Sima, Chonghao and Lu, Tong and Yu, Qiao and Dai, Jifeng},
  journal={IEEE Transactions on Pattern Analysis and Machine Intelligence},
  year={2024},
  publisher={IEEE}
}

@article{hsu2021multi,
  title={Multi-target multi-camera tracking of vehicles using metadata-aided re-id and trajectory-based camera link model},
  author={Hsu, Hung-Min and Cai, Jiarui and Wang, Yizhou and Hwang, Jenq-Neng and Kim, Kwang-Ju},
  journal={IEEE Transactions on Image Processing},
  volume={30},
  pages={5198--5210},
  year={2021},
  publisher={IEEE}
}

@misc{nvidia_modelopt,
  title        = {NVIDIA Model Optimizer Documentation},
  author       = {{NVIDIA}},
  year         = {2025},
  url          = {https://nvidia.github.io/Model-Optimizer/},
  howpublished = {\url{https://nvidia.github.io/Model-Optimizer/}},
  note         = {Accessed: 2025-12-15}
}

@misc{nvidia_tensorrt,
  title        = {NVIDIA TensorRT Documentation},
  author       = {{NVIDIA}},
  year         = {2025},
  url          = {https://docs.nvidia.com/deeplearning/tensorrt/latest/index.html},
  howpublished = {\url{https://docs.nvidia.com/deeplearning/tensorrt/latest/index.html}},
  note         = {Accessed: 2025-12-15}
}

@misc{nvidia_trtexec,
  title        = {TensorRT Command-Line Programs: trtexec},
  author       = {{NVIDIA}},
  year         = {2025},
  url          = {https://docs.nvidia.com/deeplearning/tensorrt/latest/reference/command-line-programs.html#trtexec},
  howpublished = {\url{https://docs.nvidia.com/deeplearning/tensorrt/latest/reference/command-line-programs.html#trtexec}},
  note         = {Accessed: 2025-12-15}
}

@article{WILDTRACK_Chavdarova2017TheWM,
    author = "Chavdarova, Tatjana and Baqu{\'e}, Pierre and Bouquet, St{\'e}phane and Maksai, Andrii and Jose, Cijo and Lettry, Louis and Fua, Pascal and Gool, Luc Van and Fleuret, François",
    title = "The WILDTRACK Multi-Camera Person Dataset",
    journal = "ArXiv",
    year = "2017",
    volume = "abs/1707.09299"
}

@misc{nvidia_transformer_engine_2025,
  author       = {{NVIDIA}},
  title        = {Transformer Engine},
  year         = {2025},
  howpublished = {GitHub repository},
  url          = {https://github.com/NVIDIA/TransformerEngine},
  note         = {Version v2.10 (released Dec 11, 2025), accessed 2025-12-21},
}

@article{wang2024mcblt,
  title={MCBLT: Multi-Camera Multi-Object 3D Tracking in Long Videos},
  author={Wang, Yizhou and Meinhardt, Tim and Cetintas, Orcun and Yang, Cheng-Yen and Leal-Taix{\'e}, Laura},
  journal={arXiv preprint arXiv:2412.00692},
  year={2024}
}

@article{yu2025fqpetr,
  title={FQ-PETR: Fully Quantized Position Embedding Transformation for Multi-View 3D Object Detection},
  author={Yu, Jiangyong and Shu, Changyong and Zhou, Sifan and Yu, Zichen and Hu, Xing and Chen, Yan and Yang, Dawei},
  journal={arXiv preprint arXiv:2502.15488},
  year={2025}
}

@article{luiten2021hota,
  title={HOTA: A Higher Order Metric for Evaluating Multi-Object Tracking},
  author={Luiten, Jonathon and Osep, Aljosa and Dendorfer, Patrick and Torr, Philip and Geiger, Andreas and Leal-Taix{\'e}, Laura},
  journal={International Journal of Computer Vision (IJCV)},
  year={2021}
}

@inproceedings{xiao2023smoothquant,
  title={SmoothQuant: Accurate and Efficient Post-Training Quantization for Large Language Models},
  author={Xiao, Guangxuan and Lin, Ji and Seznec, Mickael and Wu, Hao and Demouth, Julien and Han, Song},
  booktitle={International Conference on Machine Learning (ICML)},
  year={2023}
}

@article{zhou2022lowfps,
  title={APPTracker: Improving Tracking Multiple Objects in Low-Frame-Rate Videos},
  author={Zhou, Yifu and Sun, Xing and Jiang, Wenjing and Liu, Xu and Tao, Dacheng},
  journal={arXiv preprint arXiv:2208.06899},
  year={2022}
}

@article{yao2024comprehensive,
  title={A Comprehensive Survey on Post-Training Quantization for Large Language Models},
  author={Yao, Zhewei and Aminabadi, Reza Yazdani and Zhang, Minjia and Wu, Xiaoxia and Li, Conglong and He, Yuxiong},
  journal={arXiv preprint arXiv:2402.06024},
  year={2024}
}

@article{ye2021deep,
  title={Deep learning for person re-identification: A survey and outlook},
  author={Ye, Mang and Shen, Jianbing and Lin, Gaojie and Xiang, Tao and Shao, Ling and Hoi, Steven CH},
  journal={IEEE transactions on pattern analysis and machine intelligence},
  volume={44},
  number={6},
  pages={2872--2893},
  year={2021},
  publisher={IEEE}
}

@misc{wang2023exploringobjectcentrictemporalmodeling,
      title={Exploring Object-Centric Temporal Modeling for Efficient Multi-View 3D Object Detection}, 
      author={Shihao Wang and Yingfei Liu and Tiancai Wang and Ying Li and Xiangyu Zhang},
      year={2023},
      eprint={2303.11926},
      archivePrefix={arXiv},
      primaryClass={cs.CV},
      url={https://arxiv.org/abs/2303.11926}, 
}

@misc{park2022timetellnewoutlooks,
      title={Time Will Tell: New Outlooks and A Baseline for Temporal Multi-View 3D Object Detection}, 
      author={Jinhyung Park and Chenfeng Xu and Shijia Yang and Kurt Keutzer and Kris Kitani and Masayoshi Tomizuka and Wei Zhan},
      year={2022},
      eprint={2210.02443},
      archivePrefix={arXiv},
      primaryClass={cs.CV},
      url={https://arxiv.org/abs/2210.02443}, 
}

@misc{wang2026unified3dobjectperception,
      title={A Unified 3D Object Perception Framework for Real-Time Outside-In Multi-Camera Systems}, 
      author={Yizhou Wang and Sameer Pusegaonkar and Yuxing Wang and Anqi Li and Vishal Kumar and Chetan Sethi and Ganapathy Aiyer and Yun He and Kartikay Thakkar and Swapnil Rathi and Bhushan Rupde and Zheng Tang and Sujit Biswas},
      year={2026},
      eprint={2601.10819},
      archivePrefix={arXiv},
      primaryClass={cs.CV},
      url={https://arxiv.org/abs/2601.10819}, 
}

\end{document}